\documentclass[final,3p,times]{elsarticle}

\usepackage{amssymb}
\usepackage{xcolor}
\usepackage{amsmath}

\usepackage{lineno}
\usepackage{subfigure}
\usepackage{url}
\usepackage{float}

\usepackage{booktabs}
\journal{Elsevier}

\begin{document}

\begin{frontmatter}




\title{Predicting Air Quality via Multimodal AI and Satellite Imagery}


\author[l1,l2]{Andrew Rowley}

\address[l1]{Cardiff University, School of Computer Science and Informatics, Abacws, Senghennydd Road, Cardiff, CF24 4AG, UK.}

\address[l2]{Leonardo UK Ltd, 430 Coldharbour Lane, Bristol Business Park, Bristol, BS16 1EJ, UK.}

\author[l1]{Oktay Karaku\c{s}}

\begin{abstract}
One of the most important major driving forces behind climate change and environmental issues that the Earth is currently facing is the pollution of the air which remains a key symptom of negative human influence on the environment. Air pollution is often invisible to the eye which can make its detection challenging, unlike the destruction of the land or waterways. Given that air-quality monitoring stations are typically ground-based, their abilities to detect pollutant distributions are often restricted to wide areas. Satellites, however, have the potential for studying the atmosphere at large; the European Space Agency (ESA) Copernicus project satellite, “Sentinel-5P” is a newly launched satellite capable of measuring a variety of pollutant information with publicly available data outputs. This paper seeks to create a multi-modal machine learning model for predicting air-quality metrics with high precision so that it will be applicable to locations where monitoring stations do not exist. The inputs of this model will include a fusion of ground measurements and satellite data with the goal of highlighting pollutant distribution and motivating change in societal and industrial behaviours. A contemporary method for fusing satellite information with pollution measurements is studied, suggesting that simpler models can work as effectively as neural network models that are constructed with state-of-the-art architectures. A new dataset of continental European pollution monitoring station measurements is created with features including \textit{altitude, population density, environmental classification of local areas, and satellite data} from the ESA Copernicus project. This dataset is used to train a multi-modal ML model, Air Quality Network (AQNet) capable of fusing these various types of data sources to output predictions of various pollutants. These predictions are then aggregated to create an ``air-quality index" that could be used to compare air quality over different regions. Three pollutants, NO$_2$, O$_3$, and PM$_{10}$, are predicted successfully by AQNet and the network was found to be useful compared to a model only using satellite imagery. It was also found that the addition of supporting tabular data improves predictions. When testing the developed AQNet on out-of-sample data of the UK and Ireland, we obtain satisfactory estimates though on average pollution metrics were roughly overestimated by around 20\%.

\end{abstract}




\end{frontmatter}

\section{Introduction}
\label{sec:intro}
Human influence on the quality of the air we share with all life on Earth continues to contribute to its degradation. With both a global and a local perspective, accurate and reliable systems of measurement must be implemented to influence societal and industrial behaviours. Remote sensors and the statistics that they produce inform us of changes in our environment. Given advancements in deep learning, the predictive capacity of remote sensing technologies is more capable than ever. Machine learning techniques have rapidly developed in power over the past decades and have the potential to unlock new methods of analysing environmental pollution that were previously untapped. 

In this paper, multi-modal machine learning techniques are applied in order to predict air-quality indices and illustrate airborne contamination in the places we live. In response, we begin by discussing a background to pollution and remote sensing by satellite, followed by a detailed literature review on the state-of-the-art for machine learning in the remote sensing context. The proposed multi-modal ML model - \textit{AQNet} - is then presented along with several comparison studies. AQNet is also tested in out-of-distribution data of the UK and Ireland. We conclude the paper with a brief summary and sharing of insightful remarks.

\section{Background \& Related Work}\label{sec:back}
The negative impact of human beings upon the Earth manifests itself through the pollution that affects natural processes and environments wherever it acts, whether it may be: pollution of the seas and marine life; global warming; acid rain and smog; thermal pollution; groundwater poisoning; soil contamination; radioactive pollution; or even light and noise disrupting the natural world around us \cite{spellman2017science}. 

The contamination of the air can be the least visible when considering all the mediums we could pollute. Chlorofluorocarbons break down stratospheric ozone weakening our protection from solar radiation, while ozone itself within the troposphere is a key component of smog. Sulphur dioxide emissions from fossil fuel combustion poison the air we breathe and when mixed with water vapour form acids that rain down upon us. Nitrogen oxides also acidify rain and form photochemical smog – made worse by particulate matter aerosols that also may have the worse effect on the respiratory health of all airborne pollutants (See Table \ref{tab:pollutants}) \cite{spellman2017science}.

\renewcommand{\arraystretch}{0.75}
\begin{table}[ht!]
    \centering
    \begin{tabular}{p{3cm}p{6cm}p{6cm}}
\toprule 
\sc{\textbf{\large{Pollutant}}}&	\sc{\textbf{\large{Common non-natural sources}}}&	\sc{\textbf{\large{Dangers}}}\\
\toprule
Nitrogen Oxides ($NO, NO_2,\dots,NO_x$)&	Combustion of coal and gasoline&	Photochemical smog\\
&&
Acid rain as $HNO_3$\\
\hline
Carbon Oxides ($CO, CO_2$)	& Incomplete combustion of fossil fuels & Toxicity by respiration\\\hline
CFCs (Chlorofluorocarbons)	&Aerosols	&Degradation of the ozone layer\\
	&Refrigerants	&Reduced global coverage against solar radiation\\
	&Industry&\\	
\hline
Sulphur Dioxide ($SO_2$)& Fossil fuel combustion&	Respiratory system, health\\
&&		Acid rain through sulphurisation\\
&&		Visibility reduction by solution in atmosphere\\
\hline
Ozone ($O_3$)	&Reactions between $NO_2$ and Volatile Organic Compounds&Health and ecological damage\\\hline
Particulate matter ($PM_{2.5}, PM_{10}$ etc.)&Industrial processes&	Myriad respiratory conditions\\
&	Fuel combustion	&Smog\\
&	Vehicle emissions	& Catalysis of other atmospheric pollutants\\\hline
Lead particulates&	Gasoline combustion by automobiles&	Toxicity\\
&	Industrial processes&	Brain damage\\
\bottomrule
    \end{tabular}
    \caption{Common atmospheric pollutants and their dangers \cite{spellman2017science}}
    \label{tab:pollutants}
\end{table}

Additionally, the sources of air pollution are not always apparent at first glance. For example, in work collated by Harrison and Hester \cite{10.1039/9781847559654} studies at Heathrow Airport are showing that counter-intuitively ``between 5\% and 30\% of the local $NO_x$ contribution is related to aircraft, whereas the remaining 95\% to 70\% is from road traffic".

Man-made environments themselves can impact air pollution in unforeseen ways which may result in poorer air quality relative to rural areas. The “urban heat island” effect leads to increased temperatures in cities, which as well as increasing air pressure and distribution can facilitate the rate at which secondary chemical reactions may take place, releasing more pollutants than otherwise \cite{10.1039/9781847559654}.


An important challenge involved in predicting air pollutant distribution includes its transport rate across large areas. Particularly within cities, detections of pollutant concentrations measured may only provide insights at a very local level \cite{10.1039/9781847559654}. To understand what risks societies and ecosystems face, accurate measurement of such pollutants must be possible. Remote sensing provides us with these tools.

The global reach and technical development of satellites put them among the most effective tools for monitoring the Earth. Compared to remote sensing via aircraft or drones, satellites have considerable advantages. Larger areas may be studied, and more stable and clear images can be obtained with stable historical data, in addition to the most transparent reason being that aircraft are regularly required to land, high cost and weather-related flight requirements.

In general, two varieties of satellites may be found: 
(1) Geostationary: satellites are fixed relative to Earth’s surface and can monitor only 30\%-40\% of the world, (2) Polar/Near Polar: satellites sweep the Earth’s surface pole-to-pole and can scan swaths of the entire globe, but are not always within data transmission range.
In both cases, electromagnetic radiation (in general) provides the key to recording information. Each band of the electromagnetic spectrum has unique properties that can aid scientists in studying the environment. For example, visible blue light can penetrate water for approximately 10-20 m, whereas red light cannot. Vegetation reflects infrared very strongly, but the atmosphere is transparent to infrared radiation \cite{cracknell2007introduction}.

Land-use regression (LUR) and Land-use land-cover classification (LULC) are two fields of remote sensing study among which statistical analysis techniques play a key role (land-cover being an indication of the Earth’s physical characteristics and land-use referring to how these characteristics are used). Prior to the accessibility of ML techniques in the early 2010s, traditional statistical techniques were applied to LUR and LULC tasks and commonly implemented linear or logistic regressions \cite{hoek2008review}. Steps away from more traditional remote sensing analysis techniques were already being taken for the prediction of pollution, as highlighted by Young \emph{et. al.} \cite{young2016satellite}; the authors compared modelling techniques using “kriging” against those using satellite data as input when considering the prediction of pollution far from monitoring stations and found improvements versus traditional techniques.

Examples of earlier work applying ML techniques include the use of SVMs (Support Vector Machines) to identify patterns within remotely sensed images \cite{huang2002assessment}. Pollution measurement data alone had been shown to be sufficient in building accurate NNet models as shown by Alimissis \emph{et. al.} using data from neighbouring monitoring stations from Attica, Greece \cite{alimissis2018spatial}. However, advances in deep learning drew interest from researchers within the remote sensing field, particularly those with research that required image processing. 

The power of CNNs to handle multi-spectral data and convolve this to identify features \cite{ghamisi2017advances} is evident in the increase in the body of the academic literature from 3 papers in 2014 to 73 in 2016 \cite{zhu2017deep}. Sharma \emph{et. al.} \cite{sharma2017patch} developed an image-patched-based CNN approach to classifying RS imagery which they argued mitigated the lack of fine-grained structures often found within satellite images.

In 2019, LULC classification, object detection, scene recognition, segmentation, and change detection using RS imagery were the most common applications of ML in that order \cite{ma2019deep}, with CNNs being the most commonly applied tool by a large margin. Heydari and Mountrakis compared NNet efficacy to that of classical SVMs and found that transfer-learning from pre-trained CNNs, despite often having only their RGB channel parameters non-randomly allocated, perform better than SVMs unless ``rich feature maps" are extracted from datasets \cite{heydari2019meta}.

Subsequent developments have led to significant advancements in technology, for example with hyperspectral image data and LIDAR data fused together multimodally as input to neural networks \cite{yuan2020deep}. Indeed, one may observe evidence of a trend towards multimodal techniques increasing model predictivity, though key issues with this technique include a requirement for sourcing suitable multitemporal samples upon which to run the model post-development. 
Nevertheless, studies including that of Johararestani \emph{et. al.} \cite{zamani2019pm2} have demonstrated the predictive capability of multimodal satellite and ground-based models, although the authors suggest that the reduction of features by importance should be an important step in reducing model scale.

In the 2020s, according to Schneider \emph{et. al.} \cite{schneider2020satellite} traditional NN methods remain in use with an emphasis on transfer learning. In their paper, Schneider \emph{et. al.} used random forests of decision-tree algorithms to create maps of $PM_{2.5}$ distributions around Great Britain, which could inform studies on exposure to $PM_{2.5}$ matter. They conclude by highlighting the potential power of the Copernicus program in an ML/RS context. Otherwise, the field of ML in RS is ever-growing. From studies of drought \cite{fankhauser2022estimating} to water-quality \cite{uddin2022comprehensive}, the robustness of machine learning methods and their ability to handle huge streams of input data make them important tools in supplying predictive insight to environmental remote sensing problems. 

Since its launch, the Sentinel-5P (S5P) satellite has inspired the creation of much work. An early study by Guanter \emph{et. al.} in 2015 used simulated S5P data to anticipate the potential for analysing chlorophyll fluorescence, given the precursor’s status as the “first imaging spectrometer to deliver a continuous spectral sampling of red and near-infrared spectral regions” \cite{guanter2015potential}. 
Sentinel-5P’s lead instrument, TROPOMI has a resolution that provides an unprecedented level of detail to be studied. According to Efremenko and Kokhanovsky \cite{efremenko2021foundations}, this is high enough to be problematic and special care is required to retrieve the height of the pollution layers from raw sources, however, the study of tropospheric column densities is less taxing. NO$_2$ monitoring using TROPOMI 
is typical within the academic corpus; for example, S5P NO$_2$ outputs were used by researchers to show the changes in pollution over Europe brought about by stay-at-home orders during the COVID-19 outbreak \cite{virghileanu2020nitrogen}.  


In 2021, Scheibenreif released a paper with Mommert and Borth of the University of St. Gallen \cite{scheibenreif2021b} which proposed a CNN approach to predicting air pollution distribution from Sentinel-2 and Sentinel-5P imagery alone, building upon earlier work and a novel dataset created by these same authors that year \cite{scheibenreif2021a}. They later submitted an updated paper comparing their method to alternative means of prediction \cite{scheibenreif2022toward} and supplied updated code via GitHub\footnote{https://github.com/HSG-AIML/Global-NO2-Estimation}. The summation of this work demonstrated that by using a pre-trained ResNet50 backbone from a LULC model as part of a simple CNN architecture, the fusion of S2 multispectral images with mapped distributions of tropospheric $NO_2$ (from S5P) and ground truth measurements from $NO_2$ monitoring stations, acceptance criteria for the regression of $NO_2$ pollution could be met. The authors published a dataset and associated paper \cite{scheibenreif2021a} fusing Sentinel-2 and Sentinel-5P images with ground monitoring stations around Europe. This dataset (referred to as the “$NO_2$-Dataset” henceforth) consists of 3,098 entries, one for each $NO_2$ monitoring station with each sample containing:
\begin{itemize}
    \item Name and ID for each monitoring station.
    \item $NO_2$ concentration measured from said ground monitoring station that was extracted from the EEA’s open-source service, \cite{eea2022a} and aggregated by means from September 2018 to December 2020.
    \item A Sentinel-2 images
    \item A Sentinel-5P tropospheric $NO_2$ concentration images
\end{itemize}

\begin{figure}[ht!]
    \centering
    \includegraphics[width=0.8\linewidth]{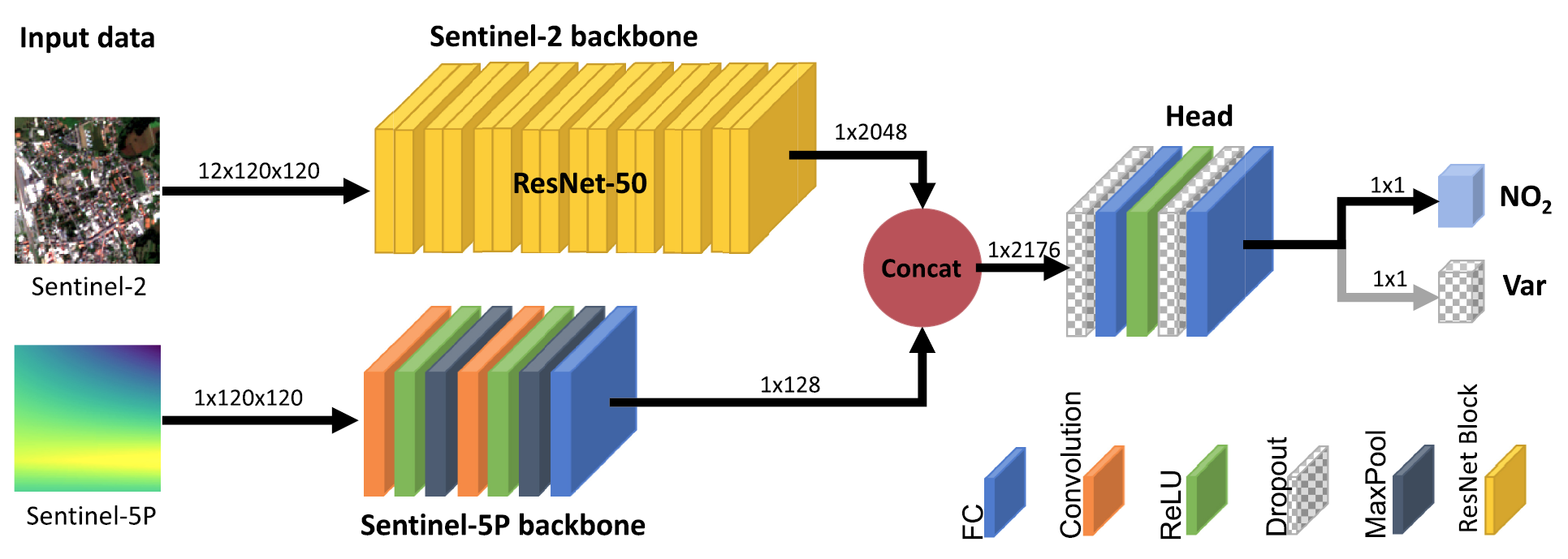}
    \caption{– The network architecture behind Scheibenreif \emph{et. al.}’s approach \cite{scheibenreif2022toward}}
    \label{fig:baseline}
\end{figure}

The aforementioned data set -  $NO_2$-Dataset - was then used as the basis from which to train their machine learning models. Sentinel-2 and 5P data from the $NO_2$-Dataset are used as input to a convolutional neural network structure (see Figure \ref{fig:baseline}) as follows:
\begin{itemize}
    \item Sentinel-2 images, size 120x120 covering 12 spectral bands are fed into a ResNet50 structure. They removed the classification layer of the ResNet50 architecture, which is leaving 2,048 neurons at the output layer.
    \item Sentinel-5P images, size 120x120 and are fed into a CNN structure with 2 convolutional and 1 fully connected layer of 128 neurons.
    \item These two input streams are concatenated, yielding a fused model with 2,176 features.
    \item These features are fed into a final “Head” layer (of 2 fully connected layers and a ReLU activation layer) that outputs the $NO_2$ regression figure.
\end{itemize}
	
Their approach obtains a 0.55$\pm$0.03 R2-Score in predicting $NO_2$ concentration over the time frame of 2018-2020. Then, the authors argue that a ResNet50 model that is pre-trained on the BigEarthNet dataset in a land-use cover classification task led to an increased model performance of 0.57$\pm$0.04. That being so, the metrics they have supplied to evidence this claim show a relatively small improvement (0.02 R2-Score). 

\section{Methodology}\label{sec:method}
\subsection{The Proposed Method - AQNet}\label{sec:aqnet}
\begin{figure}[ht!]
    \centering
    \includegraphics[width=\linewidth]{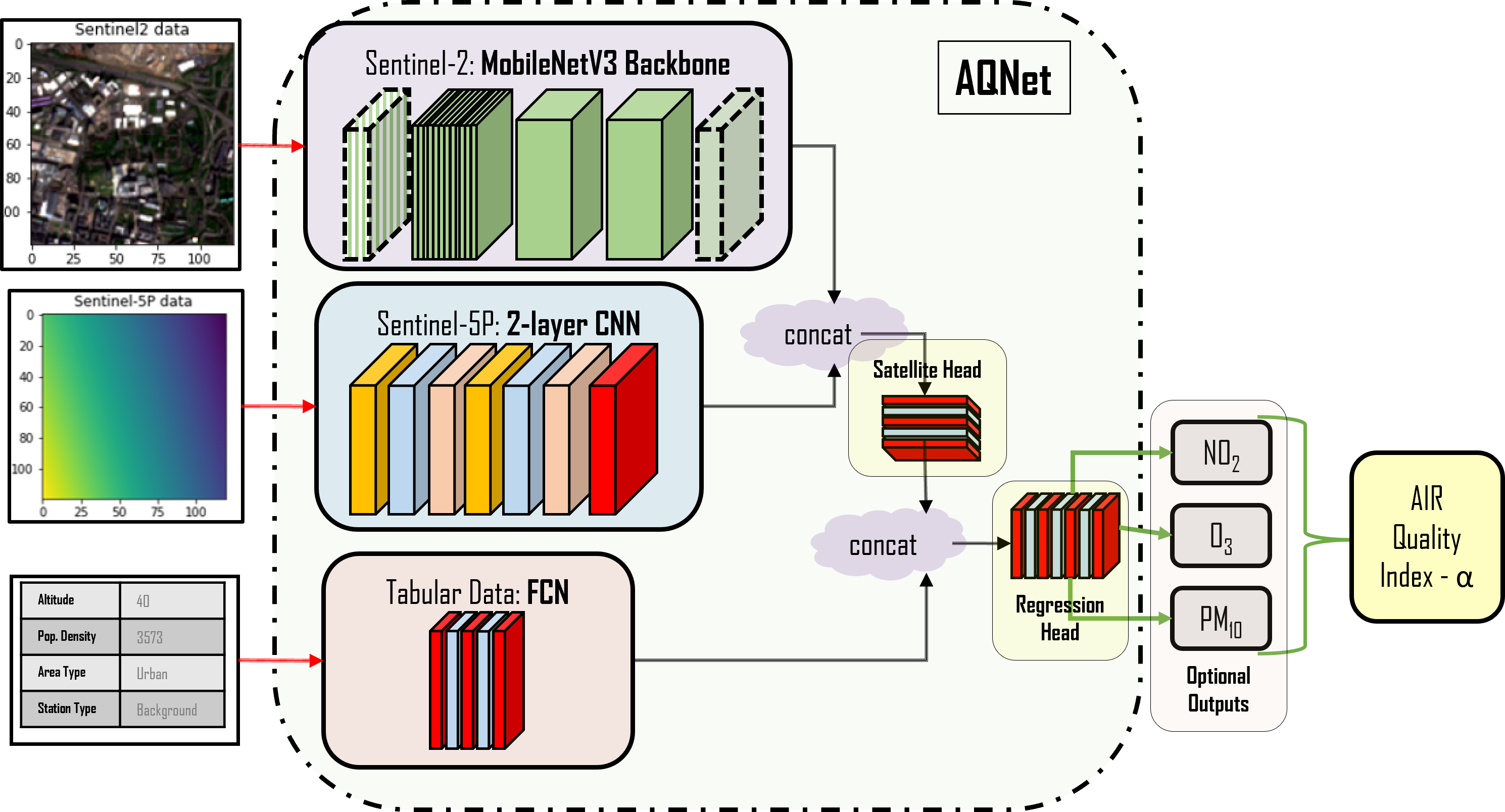}
    \caption{\textbf{AQNet} model Architecture. Named as \textbf{AQNet-single} if it returns only one of the three pollutants as output. }
    \label{fig:aqnet}
\end{figure}

In this paper, we propose a multimodal AI network that combines (1) multi-spectral satellite imagery from Sentinel-2, (2) low-resolution tropospheric $NO_2$ concentration data from Sentinel-5P satellite, and (3) tabular data that exploits some important information about the ground measurements centres, such as altitude, population density, station and area type. Our multimodal AI approach, named as \textit{AQNet}, exploits these three types of input into an optimised machine learning architecture to create predicted outputs of three important air pollutants of $NO_2$, $O_3$ and $PM_{10}$. By using the three predicted pollutants, AQNet calculated a generic air-quality metric, $\alpha$ to better quantify the predicted pollutant findings. For the purposes of more focused applications of a single pollutant concentration estimation, AQNet gives users the opportunity to predict only a single pollutant among $NO_2$, $O_3$ and $PM_{10}$ (will be named as \textit{AQNet-single} henceforth), and thus each of these outputs is named as an optional output of the model. The general architecture of the AQNet model is presented in Figure \ref{fig:aqnet}.  

\textbf{Sentinel-2 backbone:} AQNet exploits the utilisation of low-power and accurate network architecture of MobileNetV3\footnote{3$rd$ version of the mobile network architecture.} as its backbone. MobileNetV3 \cite{howard2019searching} is known to be fast in semantic segmentation-like applications, and also is capable of adapting non-linearities like swish and applying squeeze/excitation in a quantisation-friendly and efficient manner. In the initial development stages of AQNet, we experienced more than 2 times running time gain via using MobileNetV3 with $NO_2$-Dataset compared to widespread backbones such as ResNet50\footnote{Residual network consisting of 50 deep layers. ResNet18 is consisting of 18 deep layers and can be seen as a lighter version of ResNet50.} (Scheibenreif \emph{et. al.} \cite{scheibenreif2022toward} backbone), DenseNet121\footnote{Densely connected convolutional networks with 120 convolutions and 4 average pools.}\cite{huang2017densely}, and ConvNeXt\footnote{A new generation convolutional model inspired by the design of vision transformers (ViT).}\cite{liu2022convnet}.
\begin{table}[htbp]
    \centering
    \begin{tabular}{p{3.25cm}p{1.5cm}p{1.5cm}p{2.7cm}p{2.7cm}p{2.25cm}}
    \toprule
    & ResNet18 & ResNet50 & MobileNetV3 \cite{howard2019searching} &  DenseNet121 \cite{huang2017densely} & ConvNeXt \cite{liu2022convnet} \\
    \toprule
    Run time ($t$) & 4.5 hours & 9 hours & \textbf{4 hours} & 8 hours & 13 hours\\
    R2-Score ($r$) & 0.579 & 0.590 & \textbf{0.596} & 0.588 &\textbf{0.596} \\
    \hline
    Efficiency-Score ($r/t$) & 0.129 & 0.0655 & \textbf{0.1490} & 0.0735 & 0.0458\\
    \bottomrule    
    \end{tabular}
    \caption{Efficiency analysis for the backbone architectures. AQNet backbone MobileNetV3 is the fastest and best-performing architecture among others.}
    \label{tab:efficiency}
\end{table}

Each Sentinel-2 image file consists of 12 spectral bands of optical satellite imagery. During the pre-processing of the images into the NO2 dataset, images were corrected for atmospheric effects and cropped to 120$\times$120 pixel size, equivalent to 1.2km squares, each of which is centred on each ground monitoring station. Furthermore, due to having spectral resolution ranging from 10m to 60m, Sentinel-2 imagery has been upsampled and brought all band resolutions to 10m for consistency (Please see \cite{scheibenreif2021b} for more details).

\textbf{Sentinel-5P $NO_2$ backbone:} Since S5P concentration data is a single band and same size imagery format input like S2 imagery, we follow the suggestions by Scheibenreif \emph{et. al.} \cite{scheibenreif2021a} and utilised a 2-layer convolutional neural network (CNN) followed by maxpool and fully connected network (FCN) layers as for the S5P backbone. This provides us with a low-power and significantly smaller architecture as the S5P backbone. Extracted features from both satellite backbones for the S2 and S5P information are then fused in a satellite head that consists of 2 layers of FCNs.

Sentinel-5P tropospheric concentration images have been extracted from the Copernicus Open API Hub \cite{esa2022e}, with a resolution of 5x3.5km. First, pre-processing has been applied to this data to remove clouds and negative weather conditions. Then, each S5P data was mapped to a 10x10 km grid over Europe. Based on the position of each ground monitoring station a geographically near sample of this mapped data has been created matching the time from an air-quality measurement being recorded on the ground to the nearest time that the Sentinel-5P satellite overflew that station. Finally, each image has been pre-processed to yield an effective resolution of 10m.

\textbf{Tabular data backbone:} One of the most important parts of the proposed AQNet is to use tabular data along with satellite images and concentration measurements under a multimodal-AI architecture. The Tabular architecture we utilise is a basic 2-layer FCN with ReLu\footnote{ReLU stands for a rectified linear activation unit and basically returns input when it is positive, and 0.0 when negative.} activation functions. Tabular backbone takes 8 text/numerical data as its input - which are (1) altitude - numerical, (2) population density - numerical,  (3-5) binary features encoding whether the station is in a rural, suburban, or urban area, and (6-8) binary features encoding whether the station is a traffic, industrial or background monitoring station. Please note that the selection of the tabular information is not exhaustive. Depending on the data available it can be extended and richer tabular information might be created. After passing the information to the Tabular data backbone, it then creates 32 features which are fused with the satellite data in the regression head. 

\subsection{3-Pollutant Data Set}\label{sec:data}
As can be seen from Figure \ref{fig:aqnet}, the regression head is designed to create three outputs of 3 pollutants of $NO_2, O_3$ and $PM_{10}$. Within AQNet architecture all three pollutants can be predicted at the same time, along with another option to predict pollutants one by one depending on some specific needs.  In order to benefit from different pollutants' effects on each other and extract the hidden useful information, we propose using three pollutants as inputs of the architectures, and thus, 
in this section, we present a new data set for the purpose of predicting air quality index via $NO_2, O_3$ and $PM_{10}$. 3-pollutant data set aims to primarily promote pollutants Sentinel-5P mission provides. For this reason, due to their importance in air quality predictions, $NO_2$ and $O_3$ have been selected. On the other hand, we also target to promote one non-S5P pollutant. Whilst selecting it among $PM_{10}$ and $PM_{2.5}$, we tried to keep the proposed data set as big as possible. For this reason, due to the availability of the pollutants in ground stations, we added PM10 as the 3$^{rd}$ pollutant into the data set. This new dataset has 1,316 entries (excluding British and Irish samples) from distinct ground stations for pollutants $NO_2$, $O_3$ and $PM_{10}$.

Acceptable levels of concentration of all various pollutants that exist across Europe \cite{eea2022a} are set based on the WHO Air Quality Guidelines in particular. We set the highest acceptable limits for what should be deemed unhealthy concentrations of three pollutants depending on the guidelines mentioned above. Average pollution concentrations in the 3-Pollutant dataset relative to acceptance thresholds are presented in Table \ref{tab:avgthr}.
\begin{table}[htbp]
    \centering
    \begin{tabular}{p{4cm}p{4cm}p{4cm}p{2.7cm}}
    \toprule
        & $NO_2$ & $O_3$ & $PM_{10}$ \\
        \toprule
         \sc{\textbf{\large{Average}}} & \textcolor{red}{17.39}& \textcolor{teal}{54.72}& \textcolor{red}{21.30}\\
         \sc{\textbf{\large{Threshold}}}& 10.00& 60.00& 15.00\\
         \bottomrule
    \end{tabular}
    \caption{Average vs. Threshold Concentrations in $\mu g/m^3$}
    \label{tab:avgthr}
\end{table}

In the context of air quality, it should be noted that mean $NO_2$ and $PM_{10}$ figures within the dataset already exceed our thresholds for acceptable air health. This may have an effect that models trained on the dataset may be found to overpredict poor air quality. However, given thresholds were derived from WHO targets, it is unsurprising that average pollution scores could exceed them.

The distribution of stations depending on their countries in the 3-pollutant data set appears relatively diverse which should provide more resilience to models trained thereon ran upon new sources. Spain, Italy and Germany are the highest contributors to the data set with 200+ stations, whilst Portugal, the Netherlands, Poland and Romania increase station diversity on European land with 35+ stations each.

\begin{figure}[ht!]
    \centering
    \includegraphics[width=\linewidth]{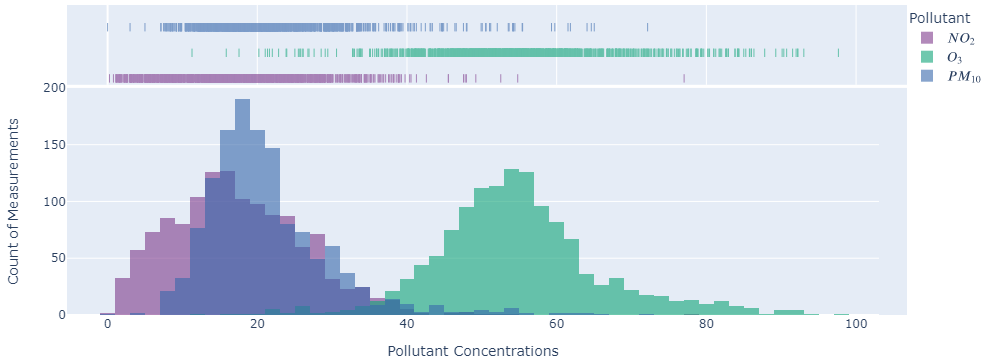}
    \caption{Histograms of pollutant concentration distribution for the three pollutants.}
    \label{fig:poldist}
\end{figure}

When we examine the distribution of each pollutant presented in Figure \ref{fig:poldist}, we conclude that (1) the distribution of ozone is normally distributed and centred about 50µg/m3, (2) the distribution of $NO_2$ and $PM_{10}$ are both positively skewed about central tendencies of 18µg/m3 and 20µg/m3, respectively, (3) the $PM_{10}$ distribution has high kurtosis, which may bias a predictive model if this distribution is not truly reflective of the global distribution of $PM_{10}$.

\begin{figure}[ht!]
    \centering
    \subfigure[]{\includegraphics[width=0.48\linewidth]{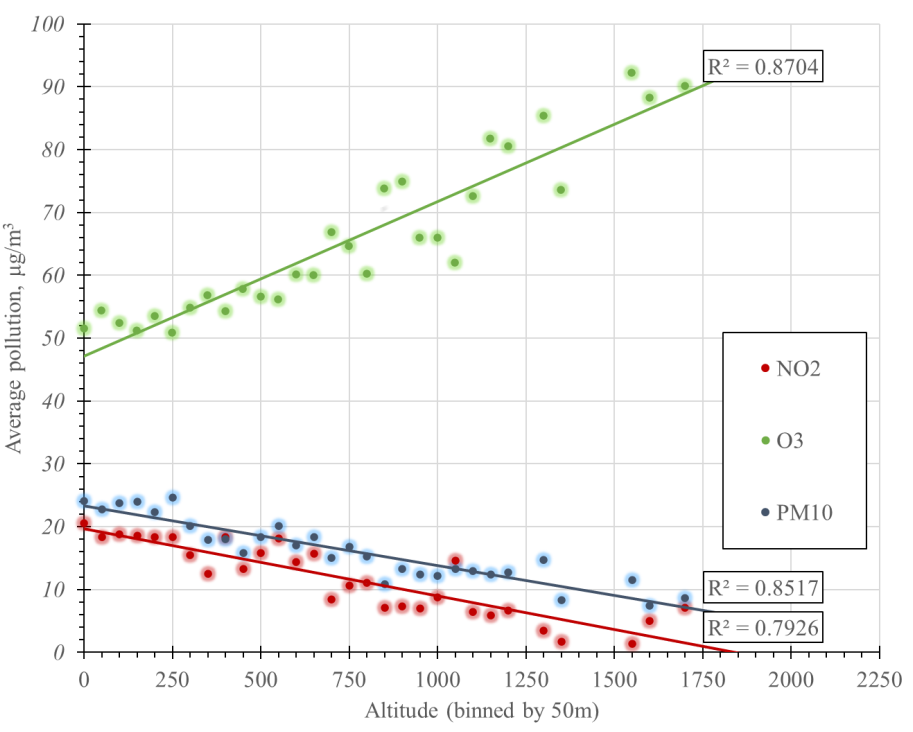}}
    \subfigure[]{\includegraphics[width=0.48\linewidth]{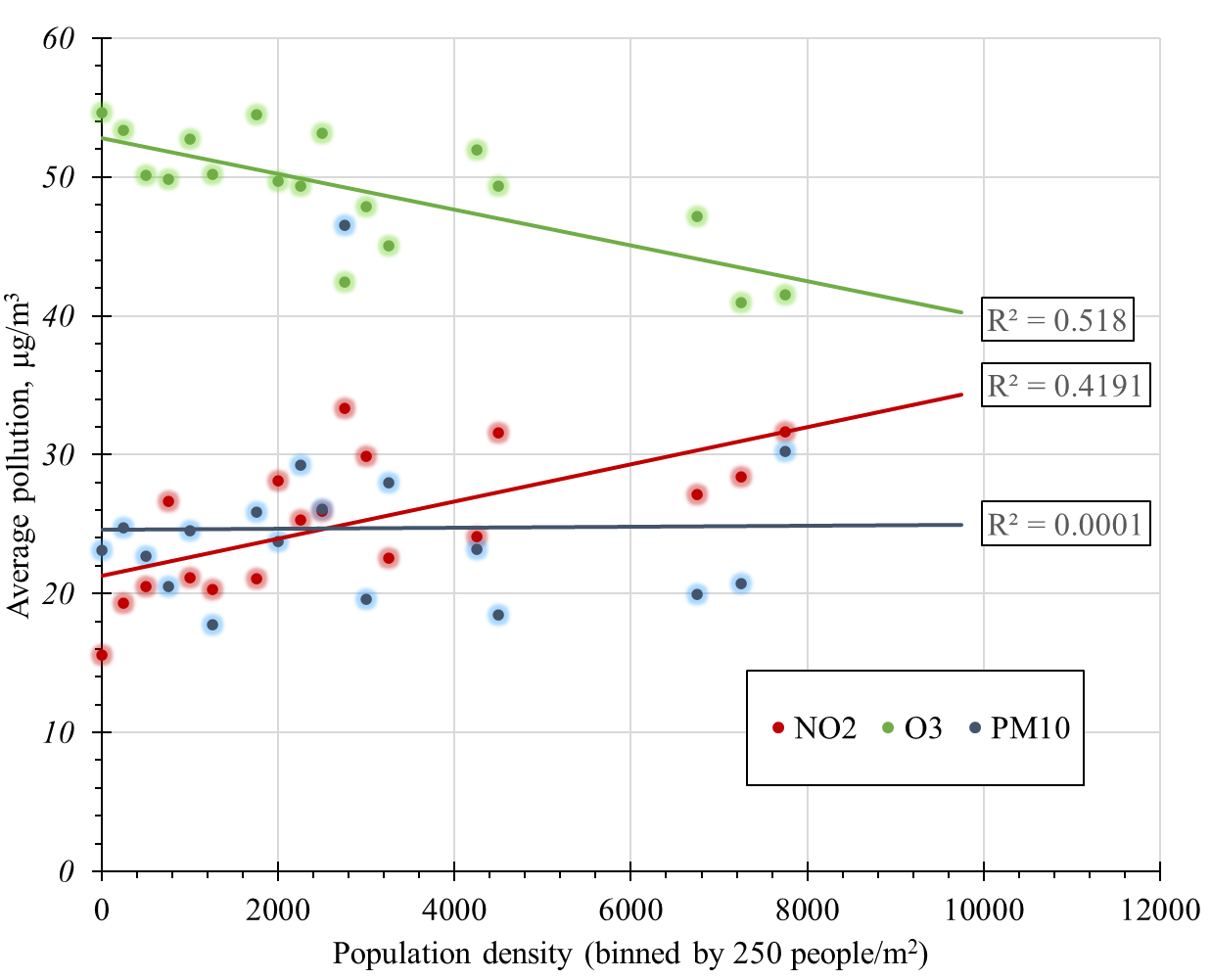}}
    \caption{(a) Altitude vs average pollution concentration, for each pollutant. (b) Population density vs average pollution concentration, for each pollutant.}
    \label{fig:corr}
\end{figure}

The two numeric features in the 3-pollutant dataset, which are "Altitude" and "PopDense" appear to have predictive power. After mean-aggregating the altitude feature to 50m bins (see Figure \ref{fig:corr}-(a)) and population density to 250 person/m2 bins (see Figure \ref{fig:corr}-(b)), we observe (1) there appear to be clear trends relating altitude to pollutant concentration, (2) in general, as altitude increases the average concentration of ozone also increases sharply, (3) as altitude increases, average concentrations of $NO_2$ and $PM_{10}$ may be expected to decrease towards zero, (4) on the other hand, increases in population density are correlated with average $O_3$ concentrations decreasing, (5) there is evidence that as population density rises, so do concentrations of $NO_2$, and (6) average $PM_{10}$ concentration and population density do not appear to be correlated in the 3-Pollutant data set.

\begin{figure}[htbp]
    \centering\includegraphics[width=0.9\linewidth]{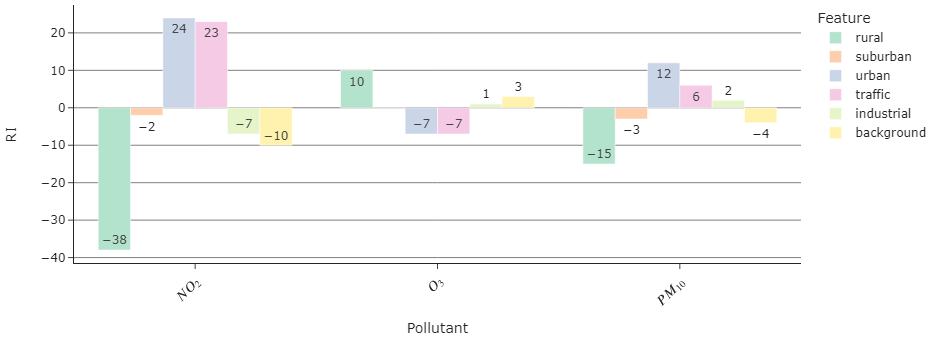}
    \caption{Relative influence upon average pollutant concentrations for each binary feature and pollutant within the dataset.}
    \label{fig:ri}
\end{figure}

As of the last analysis of the 3-pollutant data set, in considering the predictive power of each binary feature:
\begin{align*}
    \text{Area Type} &= \{rural, urban, suburban\}\\
    \text{Station Type} &= \{traffic, background, industrial\},
\end{align*}
many such features seem to influence pollutant concentration. In order to analyse this, we calculate the following statistic named ``relative influence"
\begin{align}
\mathcal{R}\mathcal{I} = \dfrac{\phi_1 - \phi_0}{\phi_0 + \phi_1}
\end{align}
where $\phi_1$ and $\phi_0$ denote average pollutant concentrations when a feature is true or false, respectively. In Figure \ref{fig:ri}, we compare this “influence” statistic across all 6 features and 3 pollutants. Examining Figure \ref{fig:ri}, we observe that stations in rural settings have decreased $NO_2$ and $PM_{10}$, and increased $O_3$ concentrations. Furthermore, Urban and traffic features, however, have a strong influence for increased levels of $NO_2$ and $PM_{10}$ when "true" whilst being situated in suburban areas does not appear to have a strong influence on average pollutant concentrations. In terms of each pollutant: (1) $O_3$ concentrations being high for rural is true, and low when urban and traffic features are false. (2) $NO_2$ concentrations are low when rural, industrial and background features are true, and high when urban and traffic features are true. (3) $PM_{10}$ concentrations being low is most influenced by the rural feature being true, and highest when the urban feature is true.


\subsection{Air Quality Index}
To ascertain the efficacy of the developed model under 3-pollutant data set, we define an air-quality index. Our original motivation in carrying out this work for multiple pollutant predictions and the proposal of the 3-pollutant data set is to prototype a tool for predicting the air quality of any region of interest. The revised dataset of 3 pollutants will be enough to illustrate the idea behind the creation of such an index, based upon the pollutant threshold values given in Table \ref{tab:avgthr}. 

Under the consideration that each pollutant's contribution to air pollution is the same when higher than the threshold, we define our generic air-quality index of $\alpha$ as
\begin{align}\label{equ:alpha}
\alpha = 1 - \dfrac{1}{N}\sum_{k=1}^{N} \dfrac{Th_k - p_k}{Th_k}
\end{align}
where
\begin{itemize}
    \item $\alpha$ denotes the calculated "air-quality".
    \item $N$ is the number of pollutants to be aggregated.
    \item $Th_k$ is the threshold for the $k^{th}$ pollutant to be deemed at an unhealthy level.
    \item $p_k$ is the predicted concentration of the $k^{th}$ pollutant outputted by the model.
\end{itemize}
The air quality metric defined in (\ref{equ:alpha}) will create a metric ranging from 0 to $\infty$. We quantify these values as given below:
\begin{align} \text{Air Quality Quantification Steps} = 
    \begin{cases}
      \alpha = 0 & \text{no pollution predicted,} \\
      0 \leq \alpha \leq 1 & \text{presence of some pollution,} \\
      \alpha > 1 & \text{unhealthy levels of pollution.} 
\end{cases}
\end{align}

\section{Experimental Analysis}\label{sec:exp}
We tested our developed approach, AQNet, under two experimental analyses: (1) model evaluation - compares the proposed approach to the baseline models from Scheibenreif \emph{et. al.} \cite{scheibenreif2022toward}. (2) out-of-distribution test - a selection of 36 monitoring stations across Britain and Ireland are used to assess AQNet performance. 

\subsection{Model Setup and Training}

The predicted pollutant concentrations are average values for each ground location over the entire period of 2018–2020. The model was trained by using the average pollutant measurements as ground truth with S2, 25P and tabular inputs and optimised to predict the average concentration. The time period utilised in this study follows the data set initially created by the baseline model in \cite{scheibenreif2021b}. Please note that whilst creating the dataset, no relevance to the pollutant or any health concern is taken into account, and please see \cite{scheibenreif2021b} for the details.

Particularly, all the AQNet architectures utilise a dataset of 1,318 monitoring stations with S2 12-band images, S5P, tropospheric NO2 density images, altitude of the station, population density, True/False information whether the station is in a rural, suburban, or urban area, and True/False information whether the station is a traffic, industrial or background monitoring station. The data set was randomly split via 2:1:1 (660 - 329 - 329) partitions as train, validation,  and test samples, respectively. The batch size of the model was set to 40 (to ensure no batch would contain 1 data point). The number of epochs was set to 30 with early stopping possible after 25 epochs whilst the learning rate was 0.001 and no drop-out was utilised. The settings above have been obtained after an optimisation stage and assumed to be the best possible selection after this analysis. The presented results in the following sections are an average of 10 runs in order to establish reliable statistics comparable to the baseline in \cite{scheibenreif2021b}.

\subsection{Model Evaluation}
In the first set of experiments in this study, we test the performance of the proposed method - AQNet, under the newly developed and revised 3-pollutant data set in predicting $NO_2$ concentrations. In addition to (1) AQNet-single (1-pollutant output), and (2) AQNet (3-pollutant output), we also tested AQNet-single architecture without tabular data in order to see the effect of tabular data addition in the 3-pollutants data set. This architecture will be named as \textbf{AQNet (No Tabular)} henceforth. 

In terms of the comparative analysis, we compared our three AQNet architectures with baseline models presented in \cite{scheibenreif2022toward}, which are (1) Satellite model: exploiting the usage of S2 and S5P imagery for the purpose of predicting $NO_2$, (2) Local: a non-satellite ML approach to predict $NO_2$ via exploiting metadata provided by the EEA such as area type, population density, and (3) Open Street Map: also a non-satellite ML approach predicting $NO_2$ by using open-sourced data including infrastructure density, etc.

\begin{figure}[htbp]
    \includegraphics[width=\linewidth]{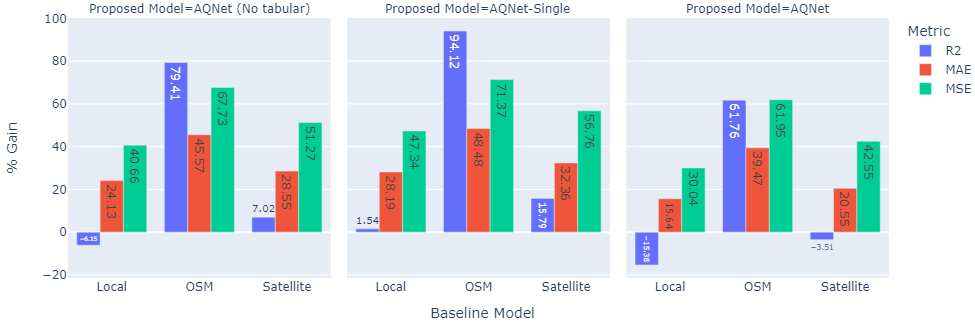}
    \caption{Percentage performance increase/decrease of Proposed AQNet architectures compared to the Baseline models of Scheibenreif \emph{et. al.} \cite{scheibenreif2022toward}. From left to right, bar-plots belong to the proposed model of {\sc{\textbf{AQNet}} (No Tabular), \sc{\textbf{AQNet-single}}} and {\sc{\textbf{AQNet}}}, respectively. }
    \label{fig:gain}
\end{figure}

For the purpose of the first set of experiments, we presented performance comparison results in Table \ref{tab:results}. Examining the results in Table Table \ref{tab:results}, by developing a new ML approach of AQNet, we managed to improve $NO_2$ concentration estimation performance compared to the baseline models of Scheibenreif \emph{et. al.} \cite{scheibenreif2022toward}.
Reiterating key metrics, we see an improvement of approximately 0.1 with AQNet-single architecture to mean R2-score metrics - albeit over differing sample sizes and restricted training data. Furthermore, 3-pollutant-output AQNet architecture successfully predicted $O_3$ and $PM_{10}$ in addition to $NO_2$ whilst having a similar R-2 score compared to satellite model\cite{scheibenreif2022toward}. It should also be noted that our simplest AQNet architecture without tabular data performed better than baseline models and showed the success of the improved architecture in AQNet even without tabular data. For simplicity, we kept AQNet architectures only pre-trained with the ImageNet data set instead of BigEarth (LULC) data. For future versions of AQNet, this dataset can be exploited with a similar potential performance increase. Figure \ref{fig:gain} depicts the percentage performance gain of the proposed AQNet architectures compared to the baseline models of Scheibenreif \emph{et. al.} 

\begin{table}[ht!]
    \centering
    \begin{tabular}{lccccc}
    \toprule
    \sc{\textbf{\large{Model}}} & \sc{\textbf{\large{Observation}}} & \sc{\textbf{\large{Pre-training}}} & \sc{\textbf{\large{R2-Score}}} & \sc{\textbf{\large{MAE}}} & \sc{\textbf{\large{MSE}}}\\
    \toprule
    \sc{\textbf{\large{Local}}}\cite{scheibenreif2022toward} & 3K & -- & 0.65$\pm$0.02 & 5.18$\pm$0.16 & 48.01$\pm$3.42\\
    \sc{\textbf{\large{Open Street Map}}}\cite{scheibenreif2022toward} & 3K & -- & 0.34$\pm$0.03 & 7.22$\pm$0.14 & 88.29$\pm$4.07\\
    \sc{\textbf{\large{Satellite}}}\cite{scheibenreif2022toward} & 3.1K & BigEarth (LULC) & 0.57$\pm$0.04 & 5.50$\pm$0.14 & 58.47$\pm$3.32\\
    \hline
    \sc{\textbf{\large{AQNet}}} (No Tabular) & 1.3K & ImageNet & 0.61$\pm$0.05 & 3.93$\pm$0.28 & 28.49$\pm$4.76\\
    \sc{\textbf{\large{AQNet}}} & 1.3K & ImageNet & 0.55$\pm$0.07 & 4.37$\pm$0.34 & 33.59$\pm$5.19\\ 
    \sc{\textbf{\large{AQNet-single}}} & 1.3K & ImageNet & \textbf{0.66$\pm$0.06} & \textbf{3.72$\pm$0.34} & \textbf{25.28$\pm$4.98}\\
    \bottomrule	

    \end{tabular}
    \caption{$NO_2$ concentration estimation results of AQNet in comparison to Scheibenreif \emph{et. al.} \cite{scheibenreif2022toward} models.}
    \label{tab:results}
\end{table}

\subsection{Out-of-distribution Analysis}
In this second set of experiments in this paper, we utilised a selection of 36 monitoring stations across Great Britain and Ireland to assess the proposed model's out-of-distribution performance. Using the multimodal-AI AQNet model trained on continental European monitoring stations, we now measure the performance of the British and Irish locations and analysed them in detail.

\begin{figure}[ht!]
    \centering
    \subfigure[]{\includegraphics[width=0.48\linewidth]{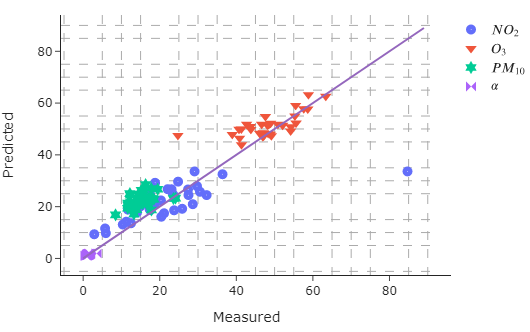}}
    \subfigure[]{\includegraphics[width=0.48\linewidth]{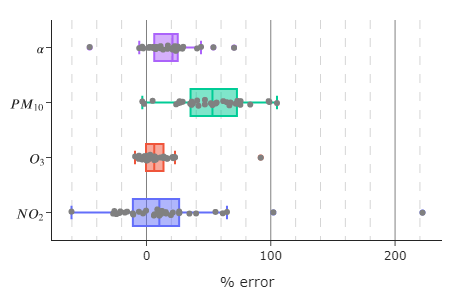}}\\
    \subfigure[]{\includegraphics[width=0.24\linewidth]{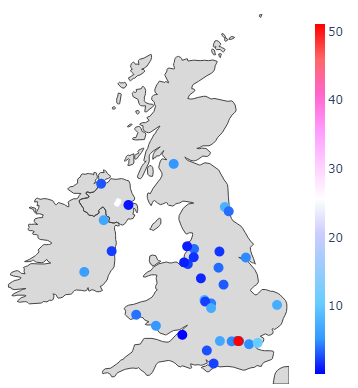}}
    \subfigure[]{\includegraphics[width=0.24\linewidth]{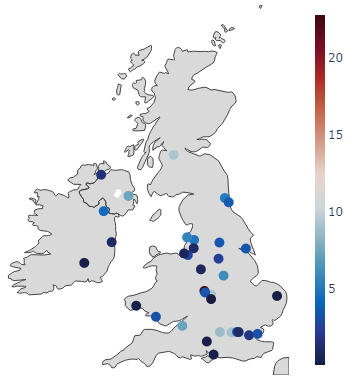}}
    \subfigure[]{\includegraphics[width=0.24\linewidth]{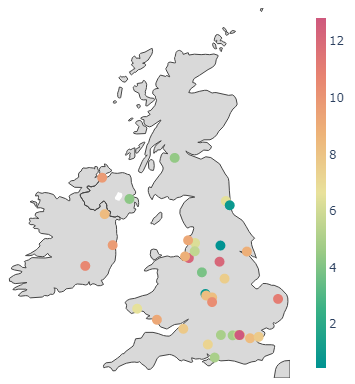}}
    \subfigure[]{\includegraphics[width=0.24\linewidth]{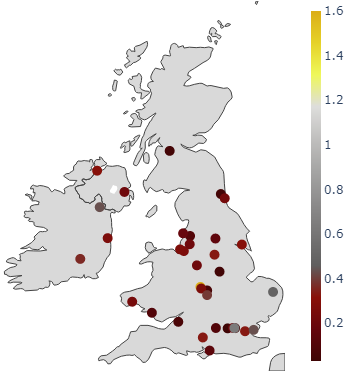}}
    \caption{(a) Regression plots of true and predicted values of each pollutant and air quality - $\alpha$. (b) Boxplots of \% error between the measured and predicted values of each pollutant. (c)-(f) Absolute error scatter maps for $NO_2, O_3, PM_{10}$ and air quality metric - $\alpha$, respectively.}
    \label{fig:ood}
\end{figure}

The collected predicted concentration results for each pollutant are then passed to the air quality metric calculation stage. Based on the equation in (\ref{equ:alpha}), Predicted-$\alpha$ (the air-quality metric) was calculated for each British and Irish pollution monitoring station. these predicted values are then compared to the "true" $\alpha$ score based on their average $NO_2, O_3$ and $PM_{10}$ measurements. When plotting the true values of alpha versus their predicted values in Figure \ref{fig:ood}-(a) and percentage error in Figure \ref{fig:ood}-(b), we observe that although the model clearly has the predictive capability, in general, $\alpha$ is overestimated. Median $\alpha$ being greater than zero suggests that the model is calibrated to detect too much pollution. An interquartile range of 18\% describes the shape of the error distribution as tight, hence our predictions may be considered to be precise but not accurate enough.

In Figure \ref{fig:ood} (c) to (f), we depict absolute error scatter plot maps for $NO_2, O_3, PM_{10}$ and air quality metric - $\alpha$, respectively whilst (b) presents boxplots of \% errors between the measured and predicted values of each pollutant. At first glance the model fitting was reasonably successful - fits on $NO_2$ and $O_3$ were close to parity, though $PM_{10}$ levels appeared overestimated Figure \ref{fig:ood}-(a).
However, when examining individual percentage error for each pollutant, it is clear that predictions for all pollutants are overestimated by around 20\% on average (Figure \ref{fig:ood}-(b)); $NO_2$ errors vary considerably which was not expected prior to experiments considering the quality of $NO_2$ fitting during model development. Median $O_3$ predictions are the closest of the three sets of predictions to parity. Even though $PM_{10}$ predictions look highly clustered with some over-estimation in (a) – \% errors vary in a higher margin compared to other pollutants; this is perhaps a consequence of the small value range $PM_{10}$ measurements in the dataset that make a small error in value creating a high percentage. This can also be seen in $PM_{10}$ absolute error scatter map in (e). 

From the scatter maps in Figure \ref{fig:ood} (c)-(f), we observe that most stations within the out-of-distribution dataset are found in England (28). Stations are concentrated in the Liverpool-Manchester urban area and the London/South East urban area.
Likewise, most of the stations in the OOD dataset are classed as being in urban, background environments. Also, altitude and population-density distributions are both positively skewed. Given the potential differences between continental European stations and British/Irish stations, these factors may have led the model to overestimate pollution. It should also be considered that perhaps environmental policies or simply the geographical qualities of GB and IE as island archipelagos lead to lower pollution averages compared to the continent.
Finally, overprediction may just be a corollary of a dearth of training data. 

\subsection{Limitations and Potential Improvements}
Upon completion of this work, many areas of improvement have become apparent. One key area that the modelling and dataset creation processes never considered was that of meteorology. Weather effects play a critical role in the transport of air pollutants \cite{spellman2017science}, yet the model has no knowledge of these as input. Even if pollutant measurements are averaged over two years, an average rainfall, humidity, or temperature statistic as an input may have influenced model predictivity – in Johararestani \emph{et. al.} \cite{zamani2019pm2}, it has been shown that wind speed and visibility were both highly important predictors of $PM_{2.5}$ concentration; data should have included many more features and reduced them during development processes via feature selection techniques. 

Likewise, though the model could have had access to geographical information, it would not have extended further than the sovereign state where each pollutant monitoring station was located. Though we maintain that this should not have been included as a categorical feature within the model, it is plausible that data encoded as a lat-long pair, or some other location feature could have been a good predictor of pollution. Although LULC characteristics were likely encoded within the 12 bands of Sentinel-2 data, it is possible that given the improvements to predictivity yielded by the inclusion of categorical features as inputs to the model, an explicit statement of land-cover concentrations could have improved results.Utilisation of Sentinel-2 bands to create useful indices for the purpose of air pollution might also be considered and developed indices might be used in the tabular data.

The current setup of the AQNet uses 3-year average pollutant concentrations. Considering these concentrations following seasonal/daily characteristics, shorter-term averages and predictions would provide further important public-health-related results. However, this requires a detailed study for developing new data sets for both ground and satellite sources and is currently within our research endeavours. 

A more informed metric for air quality would be worth pursuing. Only three pollutants were modelled which might unlikely be sufficient in assessing air quality. Likewise, the weighting of each being equally “unhealthy” within the metric might unlikely reflect reality and will be revised in future work. 

The utilised data set covers nearly 2.5 years of time and the last 5-6 months which overlap with the COVID-related lockdowns in Europe and the UK. Considering this period accommodates only a small portion of the utilised period, we believe that the effect of COVID-related reductions in air pollution should be limited and neglected. However, further analysis might cause an improved understanding of the air quality of the utilised time period.

The current setup in the proposed algorithm handles the direct use of column density measurements as model input. It has been shown in the literature \cite{novotny2011national} that estimation of ground-level NO$_2$ concentrations is possible with column density measurements with the help of surface-to-column ratios extracted from atmospheric models. Related to this, further usages of vertical column air quality data such as Pandoras, \url{https://pandora.gsfc.nasa.gov/}, would remove the limitation of having ground stations for the analysis. Future work would investigate the applicability of such an approach that might potentially increase the AQNet performance.

One final potential limitation, not discussed until now is the question of whether CNNs and FCNs are optimal tools for carrying out this task at all. As explained, CNNs have been the premier algorithm within machine vision over the last decade – but that does not guarantee that with the addition of additional non-image-based features a pure-neural network strategy would be ideal. Perhaps a mixed ML and Decision Tree approach may have led to a more successful prediction. The choice of this architecture is also the potential reason AQNet was not able to perform as well as the AQNet-Single. One can see that this is surprisingly counter-intuitive given the multi-task learning architectures in the literature \cite{zhang2012multi,huang2014deep,kendall2018multi,wang2018deep,li2019dbn,chen2019deep,sun2021deep}. Future releases of AQNet will surely investigate and utilise multi-task learning approaches for better performance.

\section{Conclusions}\label{sec:conc}
In this paper, we proposed a multimodal-AI architecture - AQNet - for air-quality measurement via a combination of satellite imagery information and some tabular information about the location of the interest. Particularly, we promoted a three-level input architecture taking (1) Sentinel-2 imagery, (2) Sentinel-5P $NO_2$ concentration image data, and (3) text/numerical data of altitude, population density, and some boolean variables defining the area and station type. 

Our proposed AQNet architectures have the flexibility to predict a single pollutant and/or three-pollutant of $NO_2, O_3$ and $PM_{10}$. These predictions are then fed to a generic air-quality calculation stage which gives a notion of whether the region of interest is polluted or not. In order for this air-quality indexing to work, we created and developed a new data set - the 3-pollutant data set - consisting of around 1.3K measurements of the pollutants in the Europe region. 

The considerable performance improvement of AQNet models under $NO_2$ and 3-pollutant data sets and the OOD test has shown us various future research directions as we discussed in the limitations and potential improvement section above. Ongoing work includes developing much-advanced S5P and Tabular backbones substituting the CNNs and FCNs, along with the development of a much richer data set including S5P imagery of $O_3$ and $PM_{10}$ pollutants.

\bibliographystyle{elsarticle-num} 
\bibliography{main}

\end{document}